\documentclass{article}

\usepackage{microtype}
\usepackage{graphicx}
\usepackage{subfig}
\usepackage{booktabs} 
\usepackage{tabularx}
\usepackage{amsmath}
\usepackage{amssymb}
\usepackage{textcomp}
\usepackage{bm}
\usepackage{svg}
\usepackage{multirow}
\usepackage{pbox}

\usepackage[hyphens]{url}
\usepackage{hyperref}

\usepackage[accepted]{icml2020}

\DeclareMathOperator*{\argmax}{arg\,max}

\newcommand{\vect}[1]{\mathbf{#1}}

\newenvironment{itemizesquish}{\begin{list}{\labelitemi}{\setlength{\itemsep}{-0.2em}\setlength{\labelwidth}{1.5em}\setlength{\leftmargin}{\labelwidth}\addtolength{\leftmargin}{\labelsep}}}{\end{list}}

\icmltitlerunning{Modelling Latent Skills for Multitask Language Generation}

\begin{document}

\twocolumn[
\icmltitle{Modelling Latent Skills for Multitask Language Generation}



\icmlsetsymbol{equal}{*}

\begin{icmlauthorlist}
\icmlauthor{Kris Cao}{deep}
\icmlauthor{Dani Yogatama}{deep}
\end{icmlauthorlist}

\icmlaffiliation{deep}{DeepMind, London, United Kingdom}

\icmlcorrespondingauthor{Kris Cao}{kriscao@google.com}

\icmlkeywords{multitask learning, few-shot learning, language generation, generative models}

\vskip 0.3in
]



\printAffiliationsAndNotice{}  

\begin{abstract}
We present a generative model for multitask conditional language generation. Our guiding hypothesis is that a shared set of latent skills underlies many disparate language generation tasks, and that explicitly modelling these skills in a task embedding space can help with both positive transfer across tasks and with efficient adaptation to new tasks. We instantiate this task embedding space as a latent variable in a latent variable sequence-to-sequence model. We evaluate this hypothesis by curating a series of monolingual text-to-text language generation datasets---covering a broad range of tasks and domains---and comparing the performance of models both in the multitask and few-shot regimes. We show that our latent task variable model outperforms other sequence-to-sequence baselines on average across tasks in the multitask setting. In the few-shot learning setting on an unseen test dataset (i.e., a new task), we demonstrate that model adaptation based on inference in the latent task space is more robust than standard fine-tuning based parameter adaptation and performs comparably in terms of overall performance. Finally, we examine the latent task representations learnt by our model and show that they cluster tasks in a natural way.
\end{abstract}

\section{Introduction}
\label{sec:intro}

Traditional approaches to conditional natural language generation 
consider each generation task (e.g.,
summarisation, paraphrasing, generative question answering) 
to be an independent task and build a task-specific model.
However, there are fundamental similarities between all of these tasks.
In an encoder-decoder model,
when the generated language from multiple tasks is in the same language, 
the structure of this language does not change across different tasks
and the generation module (i.e., the decoder)
should not have to relearn how to put words together 
into sentences for every new task.
When the input for all tasks is from the same language,
the model should benefit from sharing a common
language understanding module (i.e., the encoder).

Recent papers attempt to design a single model to perform
many language generation tasks
\citep{Raffel:19, Lewis:19, Radford:19}.
Such a universal model generally performs
worse on a particular task than their task-specific counterpart,
but the model is able to perform many diverse tasks.
In this paper, we endeavour to design a better 
universal language generation model.
Our motivating hypothesis is that modelling tasks
explicitly in a latent task embedding space allows
a universal model (for natural language generation on multiple tasks)
to represent a highly multimodal distribution of examples
from diverse tasks better, which ultimately
leads to better performance.\footnote{We use tasks and datasets interchangeably and consider each dataset to be its own task.}

We present a multitask generative model (\S{\ref{sec:model}}) that takes a sequence input $\boldsymbol{x}$
and outputs another sequence $\boldsymbol{y}$.
Our model is based on an encoder-decoder model that is augmented 
with a latent task space. 
Each point in the latent task space 
can be considered as a continuous representation of a ``skill'',
which is used to customise the encoder-decoder model to 
perform a specific task for a given input.
We model the latent task space as a mixture of Gaussians
where each training task (dataset) is represented by a Gaussian.
We provide the model with a weak supervision in the form of
a dataset identifier (i.e., which dataset a particular example comes from).
This results in an inductive bias that encourages
examples which form a specific dataset
to be solved with shared underlying skills.
In order to generate $\boldsymbol{y}$,
our model first samples a skill variable $\boldsymbol{z}$ 
from the mixture of Gaussians
and outputs $\boldsymbol{y}$ conditioned on both $\boldsymbol{x}$ and $\boldsymbol{z}$.
Crucially, our model is able to generate different outputs given the same
input $\boldsymbol{x}$ by changing $\boldsymbol{z}$.
We show how to train (\S{\ref{sec:training}}) and use this model for new examples (\S{\ref{sec:finetuning}}).

\begin{figure*}[ht]
    \centering
    \subfloat[\textsc{Full}]{\includegraphics[scale=0.4]{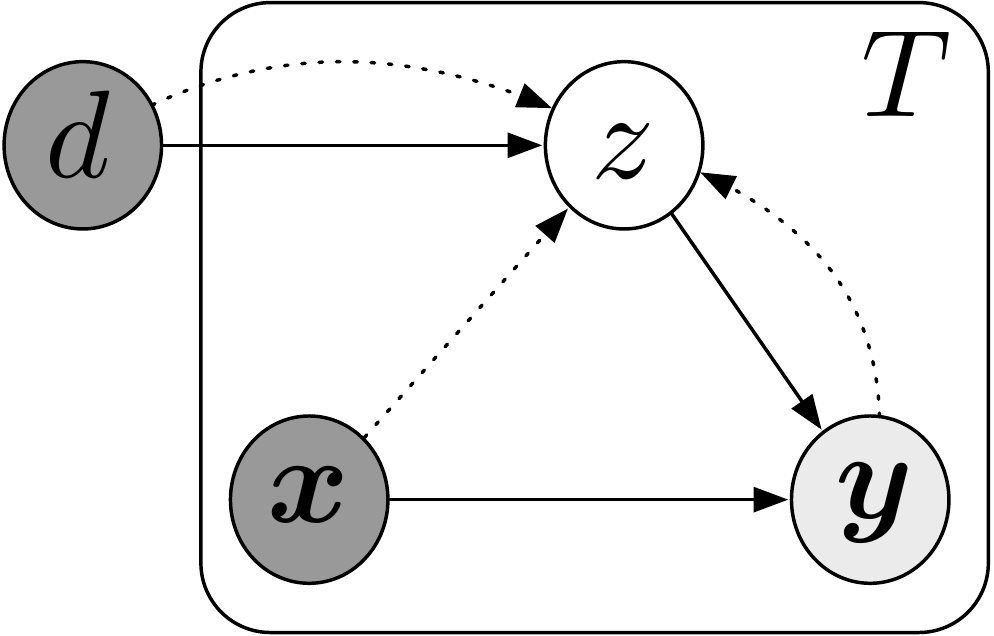}\label{fig:full_model}}
    \qquad
    \subfloat[\textsc{NoDataset}]{\includegraphics[scale=0.4]{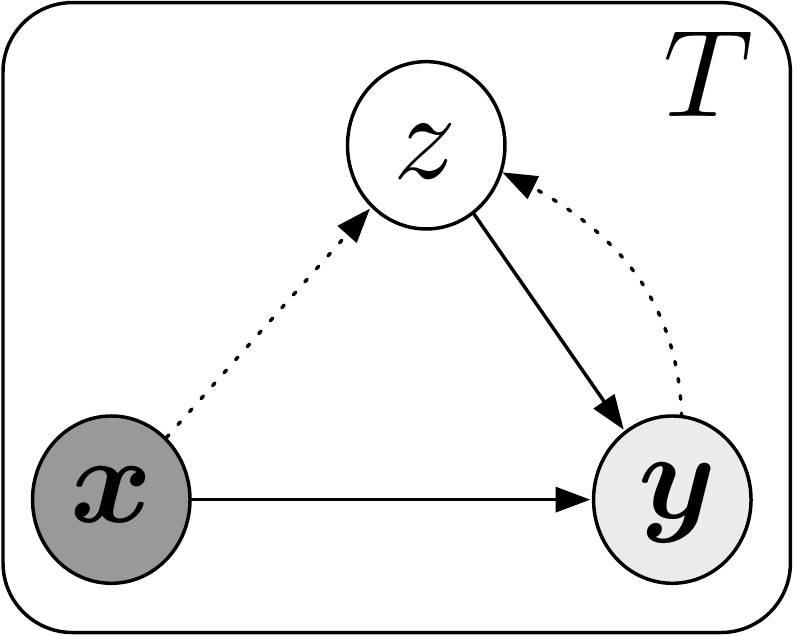}\label{fig:no_dataset_model}}
    \qquad
    \subfloat[\textsc{NoLatent}]{\includegraphics[scale=0.4]{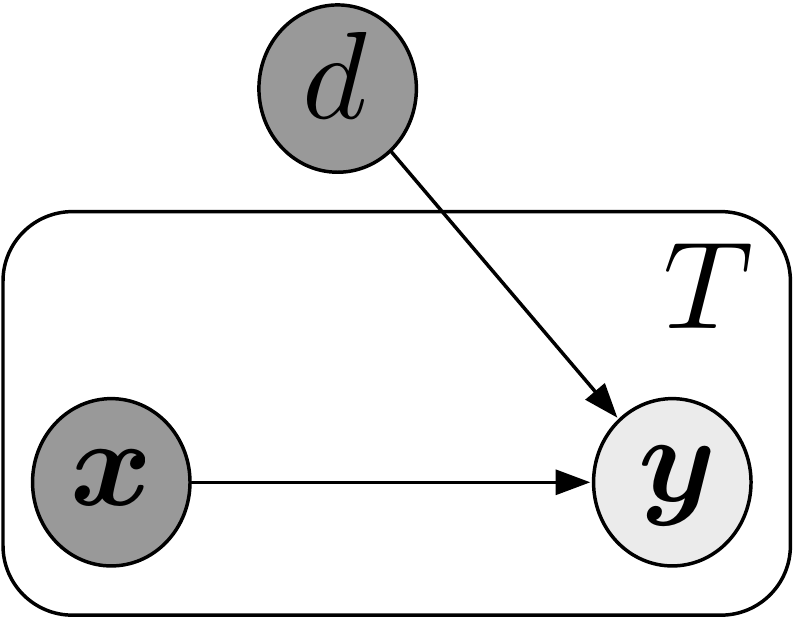}\label{fig:no_latent_model}\label{fig:base_model}}
    \qquad
    \subfloat[\textsc{Base}]{\includegraphics[scale=0.4]{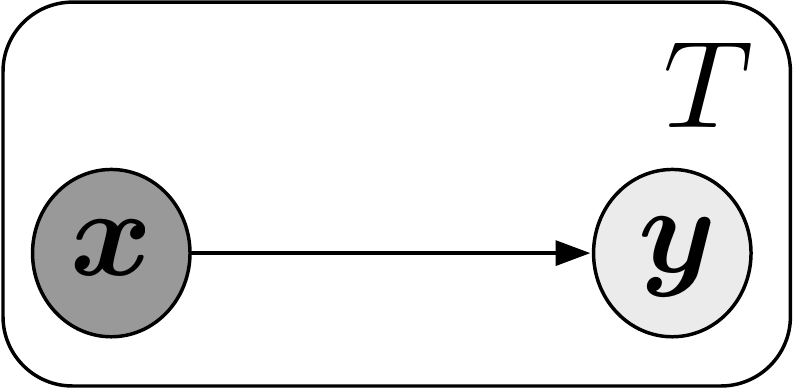}}
\caption{A graphical model depiction of our model in Figure~\ref{fig:full_model}. We also show graphical model representations of three simpler variants (Figure~\ref{fig:no_dataset_model}, \ref{fig:no_latent_model}, and \ref{fig:base_model}) that we use as baselines in our experiments. Darkly shaded variables are always observed, lightly shaded variables are observed only at training time, and others are latent.}
\vspace{-0.2cm}
    \label{fig:model}
\end{figure*}

We evaluate our model in both the multitask and few-shot settings.
We collect various datasets 
spanning across multiple tasks (e.g., summarisation, dialogue, question answering, etc.),\footnote{We limit our 
scope to English-to-English language generation tasks. We also only consider tasks where the input conditioning context is also natural language. We leave exploration on multilingual generation and incorporating image-to-text tasks to future work.}
which we discuss in detail in \S\ref{sec:datasets}.
Automated evaluation of natural language generation 
models is notoriously hard, and comparing models across 
multiple tasks is even more difficult. 
We borrow the example of GLUE \citep{Wang:18}
and report a normalised score across multiple tasks with multiple different metrics. 
We demonstrate that 
multitask learning using a latent embedding space improves 
overall performance across tasks on average---with sometimes dramatic results (\S\ref{sec:cross-task}).
In the few-shot learning setup,
we show that model adaptation based on inference in the latent task space
performs comparably to and is more robust than standard fine-tuning based parameter adaptation
(\S\ref{sec:zero-shot}). Finally, we probe the latent space and discover 
that the model clusters training tasks in a natural way
(Figure.~\ref{fig:task_mean_pca}).

\section{Model}
\label{sec:model}
Given a collection of text-to-text datasets indexed by $d \in \{1, \ldots, D\}$,
we consider the problem of generating a (natural language) output sequence 
$\boldsymbol{y}^d_t = \{y^d_{t,0}, \dots, y^i_{t,K}\}$
given a (natural language) input sequence $\boldsymbol{x}^d_t, = \{x^d_{t,0}, \dots, x^d_{t,J}\}$, where $t$ indexes an example in a dataset, and $j$ and $k$ index words in $\boldsymbol{x}$ and $\boldsymbol{y}$ respectively.
For example, for generative question-answering tasks, 
we concatenate the context and question to 
form $\boldsymbol{x}^d_t$ and predict the answer as $\boldsymbol{y}^d_t$.
For other tasks (i.e., summarisation, dialogue, and paraphrasing), we use the
input context as $\boldsymbol{x}^d_t$ and predict a summary, a continuation 
of a dialogue, and a paraphrase respectively.
Importantly, our model needs to be able
to perform different tasks (i.e., use different skills) given 
a dataset identifier $d$ and an input $\boldsymbol{x}^d_t$.
 
We present a hierarchical generative model with
a generative story as follows:
\begin{itemize}
\item Given a dataset identifier $d$ and an input example $\boldsymbol{x}^d_t$:
\begin{itemize}
\item Sample an example specific skill representation $\vect{z}^{d}_{t}$
from $p(\boldsymbol{z})$.
\item Set $y^d_{t,k=0}$ to be the start of sentence symbol.
\item While $y^d_{t,k} \neq \text{ end of sentence symbol}$:
\begin{itemize}
    \item Sample $y^d_{t,k}$ from $p(y^d_{t,k} | \boldsymbol{x}^d_t, \vect{z}^{d}_t, \boldsymbol{y}^d_{t,0:k-1})$.
\end{itemize}
\end{itemize}
\end{itemize}
Figure \ref{fig:full_model} shows a graphical model depiction of our model.
We discuss our model architecture and the training and inference methods in the following.

\subsection{Architecture}
\begin{figure*}[ht]
    \centering
    \includegraphics[width=\textwidth]{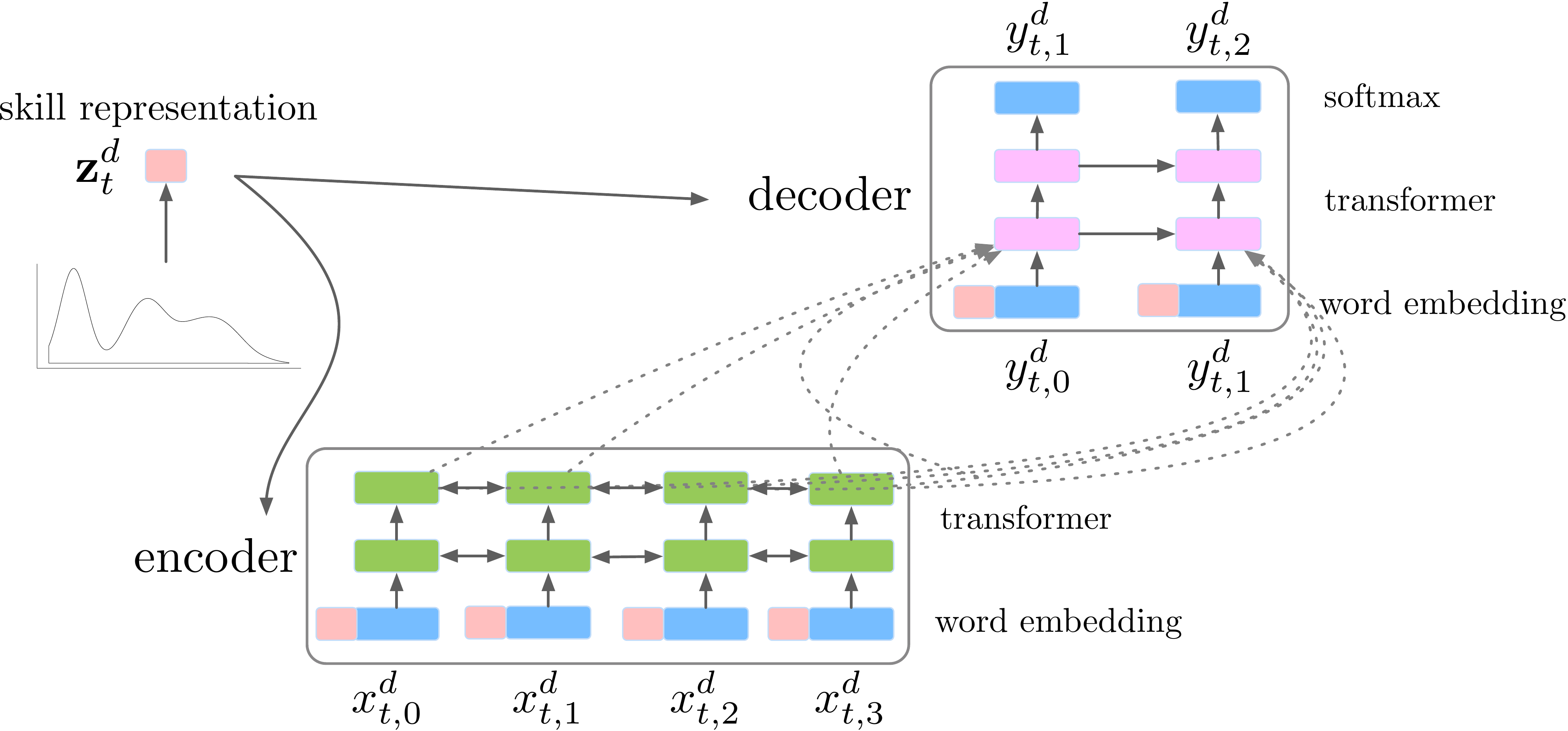}
\vspace{-0.3cm}
    \caption{An illustration of our model architecture. We sample a skill representation $\vect{z}^d_t$ and concatenate this with the embedded tokens of the context. We then pass the context through a transformer network to obtain a representation of the context. We then generate the output token by token conditioning on the input and previously generated tokens, feeding in the skill representation for previously generated tokens as well. We only show attention in the first layer of the decoder for clarity.}
    \label{fig:architecture}
\vspace{-0.2cm}
\end{figure*}
We use an encoder-decoder neural network to model $p(y^d_{t,k}  |  \boldsymbol{x}^d_t, \vect{z}^{d}_t, \boldsymbol{y}^d_{t,0:k-1})$, illustrated in Figure~\ref{fig:architecture}.

\paragraph{Encoder.}
We tokenize the input $\boldsymbol{x}^d_t$ with a SentencePiece tokenizer \citep{Kudo:18}, embed it with a word embedding layer, concatenate
a skill embedding vector $\vect{z}^d_t$ (described in detail below) to each input word,
and use a transformer encoder to obtain a sequence of contextualised input representations.

\paragraph{Decoder.} For generating outputs, 
we use the same vocabulary as the encoder input (produced by the SentencePiece tokenizer).
We reuse the encoder word embedding layer to embed words for the decoder. This word embedding layer is also used as the final softmax layer of our decoder---following previous work by \citet{Inan:16}, \citet{Press:17}, and others.
We use a decoder-specific transformer that attends to
encoded representations of the input 
and representations of previously generated outputs 
to produce a next target word $y^d_{t,k}$.
Note that the first word that the decoder conditions on is always
the start of sentence symbol,
and we concatenate the skill embedding vector $\vect{z}^d_t$ to 
the word embeddings of previously generated words when computing contextualised representations of previously generated words
using the decoder-specific transformer.


\paragraph{Latent task space.}


A na\"ive design of the latent space 
would be to assume a structureless Gaussian 
prior $p(\boldsymbol{z}) = \mathcal{N}(\vect{0}, \vect{I})$ 
and hope that examples which require similar skills to solve 
naturally cluster in this latent space. 
However, with such a unimodal distribution, we have 
little control over exactly what 
features the model will rely on to perform the clustering.

To encourage the model to cluster examples from the same dataset together,
we use a mixture of Gaussians prior:
\begin{align}
\label{eq:prior}
p(\boldsymbol{z}) = \sum_{\boldsymbol{d}\in\{1,\ldots,D+1\}} (\boldsymbol{z} | \boldsymbol{d}) p(\boldsymbol{d}), 
\end{align}
where $\boldsymbol{d}$ denotes a dataset.
Each of the Gaussians has two learnable parameters: the mean and covariance matrix
(assumed to be diagonal).
Each of the first $D$ Gaussians represents a dataset that exists in our training set.
The final $D+1$-th Gaussian is used to accommodate unseen datasets
which can be used in a certain evaluation setup (e.g., when the model is presented
with examples from a new dataset at test time).
We defer discussion of test time to \S{\ref{sec:finetuning}}.


\subsection{Training}
\label{sec:training}
The objective function that we want to maximise is:
\begin{align*}
\mathcal{L} = \sum_{d=1}^D \sum_{t=1}^T \log \int p(\boldsymbol{z}^d_t) p(\boldsymbol{y}^{d}_t  |  \boldsymbol{x}^d_t, \boldsymbol{z}^d_t) d\boldsymbol{z}^d_t.
\end{align*}
This objective function is intractable due to the integration over $\boldsymbol{z}$,
so we resort to (amortized) variational inference \citep{Kingma:13,Rezende:14,Titsias:14}.

We introduce a variational distribution $q(\boldsymbol{z}^d_t |  \boldsymbol{x}^d_t,\boldsymbol{y}^d_t)$, which takes the form of a normal distribution
$\mathcal{N}(\bm{\mu}\left(\boldsymbol{x}^d_t,\boldsymbol{y}^d_t),\bm{\Sigma}(\boldsymbol{x}^d_t,\boldsymbol{y}^d_t)\right)$.
We use a feed-forward neural network that takes as input
$\boldsymbol{x}^d_t$, $\boldsymbol{y}^d_t$ (by concatenating them as one sequence),
and a one-hot vector that represents the dataset the example comes from (i.e., a one-hot vector of the size of $D+1$)
to parametrise $\bm{\mu}$ and $\bm{\Sigma}$.\footnote{We restrict the posterior covariance matrix $\bm{\Sigma}$ to be a diagonal matrix as well.}

At training time, 
we always observe the dataset label $d$,
so $p(\boldsymbol{d} = d) = 1$ and $p(\boldsymbol{d} \neq d) = 0$
when computing the prior $p(\boldsymbol{z})$ in Eq.~\ref{eq:prior}.
However, recall that our latent task space is composed of $D+1$ Gaussians.
For 90\% of the training examples from all tasks, we keep
their original dataset label $d$.
For the remaining 10\% (chosen randomly),
we assign these examples to the $D+1$-th Gaussian
to allow the model to accommodate for unseen datasets (i.e., examples which
come from datasets that are not in the training datasets) 
at evaluation time.
The goal is to allow the the final Gaussian to
be a generalist that can be adapted quickly
to a new dataset. We describe this procedure in detail in \S\ref{sec:finetuning}. The final variational lower bound of the above likelihood function that we
optimise to train our model is:
\begin{align*}
    \mathcal{L} \geq \sum_{d=1}^D\sum_{t=1}^T \mathbb{E}_{\boldsymbol{z}^{d}_t \sim q} [&\log p(\boldsymbol{y}^{d}_t  |  \boldsymbol{x}^d_t, \boldsymbol{z}^d_t)] \\
    &- \mathbb{KL}\left[q(\boldsymbol{z}^d_t |  \boldsymbol{x}^d_t,\boldsymbol{y}^d_t) \Vert p(\boldsymbol{z}^d_t)\right].
\end{align*}
We use a single sample from $q$ to approximate the expectation and compute the 
KL between $p$ and $q$ in closed form.

\subsection{Predictions}
\label{sec:finetuning}

Recall that our model always takes a weak supervision in the form of which dataset
a test example comes from---either a particular training dataset
or a new unseen dataset.
We have three evaluation modes: in-domain evaluation, zero-shot evaluation, and few-shot adaptation.

\paragraph{In-domain evaluation.}
In this setup, we evaluate on an example from a dataset that we have seen
at training time. This is a classic multitask evaluation scenario.
Given a test example $\boldsymbol{x}^d_s$,
we use the mean of the Gaussian associated with this dataset to obtain
$\vect{z}^d_s$ and decode to generate our prediction $\boldsymbol{\hat{y}}^d_s$.
We use this evaluation mode in \S\ref{sec:cross-task}.

\paragraph{Zero-shot transfer.}
For zero-shot evaluation, the model is given an example from a new unseen dataset.
We simply use the mean of the extra $D+1$-th Gaussian to obtain
$\vect{z}^{D+1}_s$ and decode to generate our prediction $\boldsymbol{\hat{y}}^{D+1}_s$.
We use this evaluation mode to establish baseline zero-shot performance 
and compare it with few-shot learning in \S\ref{sec:finetuning}.


\paragraph{Few-shot adaptation.}
One of the crucial parts of machine learning is rapidly 
adapting an existing model to a new dataset, often with limited examples. 
In this setup, we are given a few examples from a new dataset that has not 
been seen at training time. The model must then make predictions for
more examples drawn from the same distribution.

One approach towards few-shot adaptation is to optimise 
the parameters of the model on the few-shot data, typically via gradient descent. 
This effectively retrains (fine-tunes) the model on the newly presented training data, 
and is currently the dominant paradigm for transfer learning.
However, fine-tuning all the parameters of the model 
with few examples is known to be
sensitive to the hyperparameters of the tuning procedure.
It also carries 
the risk of \textit{catastrophic forgetting} \citep{McCloskey:89,Ratcliff:90}, 
where prior knowledge learnt on the training tasks 
is not retained and performance on them dramatically decreases.



On the other hand, our model explicitly learns latent skills 
which can be used across tasks. Our approach is to 
perform inference on what latent skills 
are used in the few-shot examples, and then 
make predictions using these inferred latent skills. 
As we do not change any parameter values of our base encoder-decoder model, 
we expect this method to be more robust 
than standard gradient-based adaptation.

Specifically, given a new set of $T$ training 
examples from a new unseen dataset:
$\{\langle\boldsymbol{x}^{D+1}_t,\boldsymbol{y}^{D+1}_t\rangle\}_{t=1}^T$,
we first compute $\vect{z}^{D+1}_t = q(\boldsymbol{z}^{D+1}_t | \boldsymbol{x}^{D+1}_t,\boldsymbol{y}^{D+1}_t)$
for each new example $t$. We then compute a posterior mean of the new dataset
by averaging the posterior mean
of each new example $\vect{\hat{z}}^{D+1} :\frac{1}{T}\sum_{t=1}^T\vect{\hat{z}}^{D+1}_t$.
We finally use $\vect{\hat{z}}^{D+1}$ as the skill representation
to generate our prediction for all test examples in this new dataset.

A possible limitation of this approach is that 
our inference network $q$ might not generalise well to examples from the 
new task. 
One possible way to improve our 
estimate $\vect{\hat{z}}^{D+1}_t$
is to optimise:%
\begin{align}
\label{eq:improved}
\vect{\hat{z}}^{D+1}_t = \argmax_{\boldsymbol{z}^{D+1}_t}\log p(\boldsymbol{y}^{D+1}_t | \boldsymbol{z}^{D+1}_t,\boldsymbol{x}^{D+1}_t)p(\boldsymbol{z}^{D+1}_t),
\end{align}
using the inferred $\vect{\hat{z}}^{D+1}_t$ as the initialisation point.
This procedure improves our estimate of $\vect{\hat{z}}^{D+1}_t$ given an example $\langle\boldsymbol{x}^{D+1}_t,\boldsymbol{y}^{D+1}_t\rangle$ since $p(\boldsymbol{y}^{D+1}_t | \boldsymbol{z}^{D+1}_t, \boldsymbol{x}^{D+1}_t)p(\boldsymbol{z}^{D+1}_t) \propto p(\boldsymbol{z}^{D+1}_t |  \boldsymbol{x}^{D+1}_t,\boldsymbol{y}^{D+1}_t)$ and the proportionality constant 
does not depend on $\boldsymbol{z}^{D+1}_t$.
Furthermore, the prior
acts as a natural regulariser that
prevents our estimate from deviating too far from previously learnt values.
Similar to the above, after doing this optimisation 
for all $T$ new training examples (independently),
we then average 
$\vect{z}^{D+1}_t$ over all of the few-shot 
training examples to get $\vect{\hat{z}}^{D+1}$, which we then use to generate predictions for the few-shot test data. 

We show results with both of our few-adaptation methods---which we refer to as \textsc{Infer} and \textsc{Infer++}---and compare them with a standard fine-tuning based approach in \S\ref{sec:zero-shot}.



\section{Experiments}

\subsection{Tasks and Datasets}
\label{sec:datasets}
\begin{table}[t]
    \centering
\resizebox{0.49\textwidth}{!}{
    \begin{tabular}{l c c c c}
        \toprule
        Dataset name & \# Train & \# Test & Input len & Output len \\
        \midrule
        Gigaword & 3,803,920 & 189,651 & 31.35 & 8.23 \\
        CNN/DM & 287,226 & 13,368 & 791.38 & 55.17 \\
        NEWSROOM & 995,033 & 108,837 & 659.08 &  26.82 \\
        NYT & 137,778 & 17,222 & 995.22 & 80.48 \\
        TL;DR & 3,077,981 & 6,400 & 211.50 & 25.89 \\
        Wikihow & 168,128 & 6,000 & 508.26 & 52.35 \\
        \midrule
        MSMARCO & 502,932 & 55,578 & 67.66 & 12.89 \\
        NewsQA & 76,560 & 4,341 & 608.92 & 4.04 \\
        SQuAD & 86,821 & 5,928 & 129.86 & 3.16 \\
        \bottomrule
    \end{tabular}
    }
    \caption{Summary statistics of each of our evaluation datasets. The length of the input/output is defined as the number of whitespace-separated tokens after preprocessing.}
    \label{tab:dataset_stats}
\end{table}

Our experiments focus on monolingual text-to-text language generation. 
We collect a diverse set of tasks and datasets to evaluate our model.
We show descriptive statistics of our datasets in Table~\ref{tab:dataset_stats}
and discuss them below.

\paragraph{Summarisation.}
We use the following summarisation datasets:
\begin{itemizesquish}
    \item Gigaword news headline generation~\citep{Rush:15}. We reprocess the dataset to remove UNK tokens.
    \item CNN/Daily Mail news article summarisation~\citep{Hermann:15,See:17}
    \item NEWSROOM news article summarisation~\citep{Grusky:18}
    \item NYT news article summarisation~\citep{Durrett:16}. We use the splits provided in \citet{Xu:19}.
    \item TL;DR Reddit article summarisation~\citep{Volske:17}.
    \item Wikihow instructional article summarisation~\citep{Koupaee:18}.
\end{itemizesquish}

\paragraph{Question Answering.} We perform all question answering tasks as a language generation task, rather than span extraction. As we are primarily interested in generating correct answers, we remove all questions which are unanswerable in the MSMARCO and SQuAD datasets.
\begin{itemizesquish}
    \item MSMARCO web article QA \citep{Bajaj:16}. We concatenate all articles which have been selected by the annotator as useful as the context.
    \item NewsQA news article QA \citep{Trischler:17}. We use the consensus answer.
    \item SQuAD Wikipedia QA \citep{Rajpurkar:16}
\end{itemizesquish}

\paragraph{Others.} In the multitask scenario, we train our model on two further categories of tasks: \textbf{paraphrasing} and \textbf{dialogue}. However, we do not evaluate on them. 
We include three paraphrase corpora: ParaNMT \citep{Wieting:18}, Quora Question Pairs\footnote{\url{https://www.quora.com/q/quoradata/First-Quora-Dataset-Release-Question-Pairs}}, and MRPC \cite{Dolan:05}; we use positive paraphrase examples from paraphrase identification corpora, and generate each sentence in the paraphrase pair from the other one. We also include two dialogue corpora: the Reddit conversational corpus \citep{AlRfou:16}, and OpenSubtitles \citep{Lison:16}.

\subsection{Implementation Details}
We train all models on a cluster of 8 Nvidia V100s, with a batch size of 32 for each GPU. 
We use 6 transformer layers in the encoder and decoder, and use a word embedding size and transformer hidden layer size of 512. The dimensionality of the latent skill space was 64. 
For training, we use the Adam optimizer \citep{Kingma:14} with learning rate $3e^{-4}$, 
and optimise for $5e^{6}$ steps.
These values are chosen using the development set of each dataset; 
if a model does not have a development set, 
we take $N$ examples from the head of the training set, where $N$ is as shown in Table \ref{tab:dataset_stats}.

Our SentencePiece tokenizer is
trained on a random sample of 1 million randomly selected inputs and 
outputs from the entire set of training examples, and we keep a vocabulary of 24,000 tokens. All contexts are truncated to a maximum of 268 tokens, and all outputs are truncated to 256 tokens. For QA tasks, the question is truncated to at most 32 tokens.

For models with latent variables, we also tune the weight of the KL term (the beta parameter as in \citet{Higgins:18}) and anneal the KL term from 0 to maximum over the course of model training. We find that a beta term of 0.5 and annealing the KL term linearly over 100,000 model steps works the best across all datasets.

\subsection{Multitask Evaluation}
\label{sec:cross-task}

We first evaluate whether our model can transfer knowledge across multiple tasks without task interference. 
We compare our model---denoted by \textsc{Full}---to three baseline models which have a simpler latent variable hierarchy:
\begin{itemize}
    \item \textsc{NoDataset}: our first baseline removes the dataset index from the latent variable. Here, all examples for all datasets are generated from the same Gaussian prior with zero mean and identity covariance (see Figure \ref{fig:no_dataset_model} for a graphical model overview).
    \item \textsc{NoLatent}: the second baseline removes the latent variable $\boldsymbol{z}$, but keeps the conditioning on the dataset index. This can be viewed as collapsing each component in the latent mixture model to a point mass in the latent skill space (Figure \ref{fig:no_latent_model}). This model is analogous to a sequence-to-sequence model that is augmented with a trainable task embedding (one for each task) and is trained on multiple tasks.
    \item \textsc{Base}: the last ablation removes all conditioning, and generates each output conditioning only on the input (Figure \ref{fig:base_model}). This model is analogous to a standard sequence-to-sequence model that is trained on multiple tasks without any knowledge of the task beyond what exists in the input.
\end{itemize}
For each model, we also train it on each task independently (the single task setup, where we train a separate model for each dataset) to
assess whether the model benefits from multitask training.

\paragraph{Evaluation metrics.}
Many different metrics have been proposed for evaluating natural language generation systems. 
The most popular ones are overlap measures based on a reference, 
such as ROUGE \citep{Lin:04} and BLEU \citep{Papineni:04} for summarisation, 
and $F_1$ and exact match for question answering \citep{Rajpurkar:16}. 
However, comparing model performance across tasks and 
across metrics is difficult, 
as these metrics all take different ranges of values for different metrics and different tasks.

Motivated by the GLUE score proposed by \citet{Wang:18}, 
which aims to compare the performance of models across a range of different tasks, 
we propose a normalised score to compare multitask language generation systems. 
For each task, we first report the maximum score for 
each metric achieved across all of our models. 
We then report all model results for all metrics as percentages 
of the maximum score for that metric for that task. 
This facilitates comparison across tasks, as now the ranges 
of the metrics for each task are comparable. 
We report the best scores for each metric and each 
summarisation task in Table \ref{tab:best_results_summ}, as well as which models achieved the best score. For summarisation, the metrics we use are ROUGE-1, ROUGE-2, ROUGE-L, and BLEU. 
For question answering, we evaluate using $F_1$ only, as exact match penalises generative question answering models harshly.

\begin{table}[t]
    \centering
    \resizebox{0.5\textwidth}{!}{
    \begin{tabular}{l c c c c | c}
        \toprule
         & R1 & R2 & RL & BLEU & $F_1$ \\
        \midrule
        Gigaword & 50.14\textsuperscript{c,s} & 26.56\textsuperscript{c,s} & 47.28\textsuperscript{c,s} & 23.20\textsuperscript{c,s} & -- \\
        CNN/DM & 41.82\textsuperscript{b,s} & 16.83\textsuperscript{a,s} & 28.23\textsuperscript{b,s} & 13.67\textsuperscript{a,m} & -- \\
        Newsroom & 34.35\textsuperscript{c,s} & 24.50\textsuperscript{c,s} & 31.90\textsuperscript{c,s} & 39.45\textsuperscript{b,s} & -- \\
        NYT & 44.79\textsuperscript{a,m} & 28.32\textsuperscript{a,m} & 36.65\textsuperscript{a,m} & 30.01\textsuperscript{a,m} & -- \\
        TL;DR & 14.27\textsuperscript{a,s} & 2.14\textsuperscript{a,s} & 10.54\textsuperscript{a,s} & 2.18\textsuperscript{a,s} & -- \\
        Wikihow & 26.73\textsuperscript{a,m} & 8.13\textsuperscript{a,m} & 20.48\textsuperscript{a,m} & 7.28\textsuperscript{a,m} & --\\
        \midrule
        MSMARCO & -- & -- & -- & -- & 65.46\textsuperscript{a,s} \\
        NewsQA & -- & -- & -- & -- & 49.00\textsuperscript{c,m} \\
        SQuAD & -- & -- & -- & -- & 73.81\textsuperscript{a,m} \\
        \bottomrule
    \end{tabular}
    }
    \caption{The best results for each task and each metric. (a, b, c, d) refer to which model achieves that score in the notation of Figure~\ref{fig:model}, which are (\textsc{Full}, \textsc{NoDataset}, \textsc{NoLatent}, \textsc{Base}) respectively. (s, m) refer to whether the model was trained in the single or multitask setting respectively. These serve as the normalisation constants for the scores we report in Table~\ref{tab:multi_task}. We show the full results for all models in Appendix~\ref{app:full_results}. Current state-of-the-art R1 for CNN/DM is 44.14 \citep{Zhang:19}; Newsroom is 39.91 \citep{Shi:19} and NYT is 45.50 \citep{Xu:19}. Our model performances are in the range of these state-of-the-art numbers.}
    \label{tab:best_results_summ}
\vspace{-0.1cm}
\end{table}

\begin{table*}[t]
    \centering
    \small
    \begin{tabular}{l cccc cccc}
        \toprule
         & \multicolumn{4}{c}{Single task} & \multicolumn{4}{c}{Multitask} \\
         \cmidrule(r){2-5} \cmidrule(l){6-9}
        Dataset & \textsc{Full} & \textsc{NoDataset} & \textsc{NoLatent} & \textsc{Base} & \textsc{Full} & \textsc{NoDataset} & \textsc{NoLatent} & \textsc{Base} \\
        \midrule 
        Gigaword & 90.04 & 94.86 & \textbf{100.00} & 99.30 & 87.57 & 74.23 & 91.01 & 90.96 \\
        CNN/DM & 98.21 & \textbf{99.74} & 87.81 & 86.72 & 98.21 & 89.43 & 85.31 & 91.53 \\
        Newsroom & 83.40 & \textbf{97.95} & 97.92 & 79.13 & 92.77 & 74.85 & 80.47 & 82.36 \\
        NYT & 98.96 & 98.47 & 91.21 & 93.10 & \textbf{100.00} & 82.92 & 94.72 & 96.76 \\
        TL;DR & \textbf{100.00} & 94.47 & 83.02 & 57.97 & 87.54 & 73.08 & 61.52 & 76.16 \\
        Wikihow & 88.25 & 88.50 & 58.41 & 67.64 & \textbf{100.00} & 85.10 & 63.95 & 63.21 \\
        \midrule
        MSMARCO & \textbf{100.00} & 91.44 & 79.40 & 84.73 & 94.55 & 82.59 & 94.53 & 98.74 \\
        NewsQA & 22.57 & 25.55 & 18.08 & 21.43 & 99.38 & 86.69 & \textbf{100.00} & 98.72 \\
        SQuAD & 18.58 & 17.91 & 12.36 & 13.02 & \textbf{100.00} & 89.67 & 98.98 & 99.26 \\
        \midrule
        Average (summ.) & 93.14 & 95.67 & 86.39 & 80.64 & \textbf{94.35} & 79.94 & 79.50 & 83.50 \\
        Average (QA) & 47.05 & 44.97 & 36.61 & 39.73 & 97.98 & 86.32 & 97.84 & \textbf{98.91} \\
        Average (all) & 77.78 & 78.77 & 69.80 & 67.01 & \textbf{95.56} & 82.06 & 85.61 & 88.63 \\
        \bottomrule
    \end{tabular}
    \caption{Single and multitask results for our evaluation datasets, reported using the aggregate metric described in \S\ref{sec:cross-task}. Note that our full model results for summarisation improve with multitask training, compared to the ablated models. Multitask training seems to be uniformly beneficial for QA tasks across all models. }
\vspace{-0.2cm}
    \label{tab:multi_task}
\end{table*}

\begin{figure}[t]
    \centering
    \includegraphics[width=0.47\textwidth]{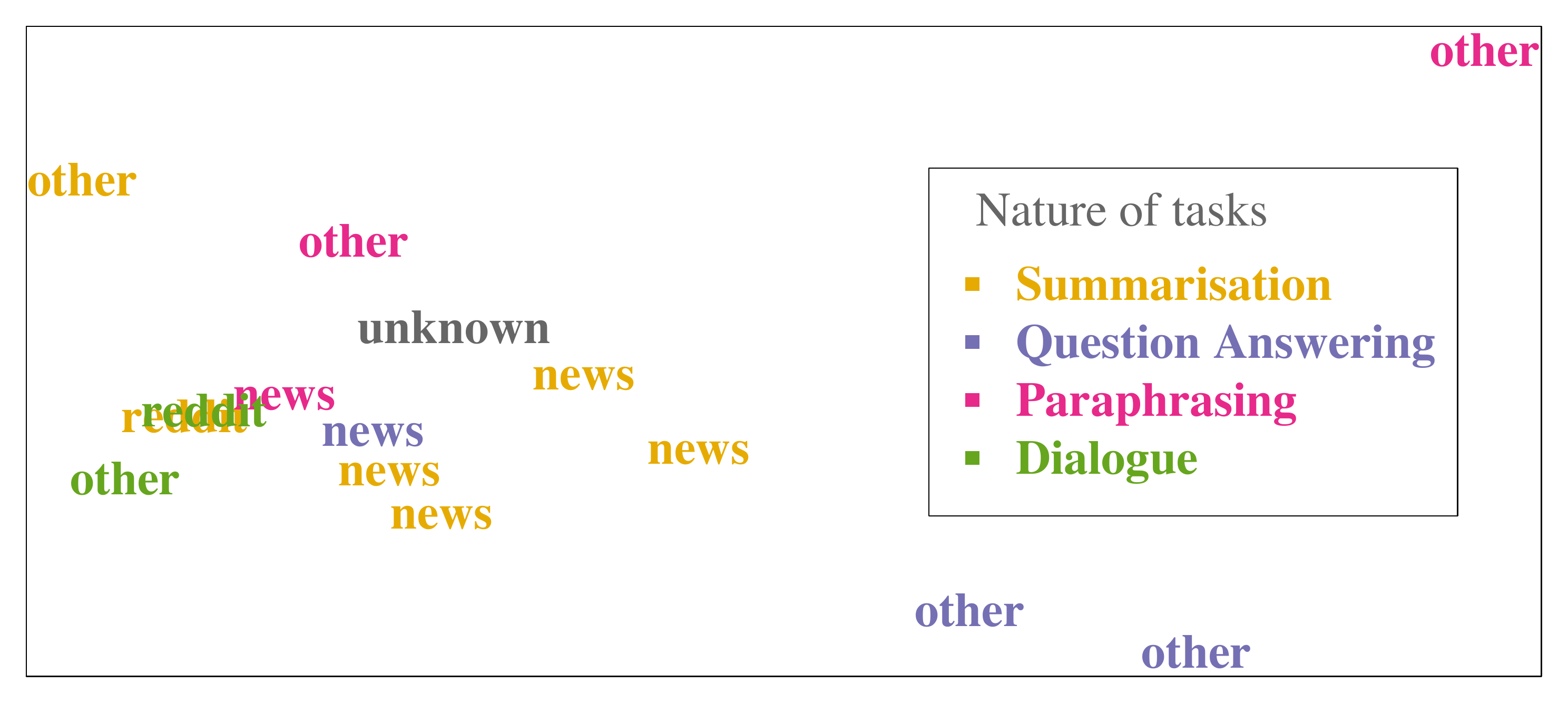}
    \caption{PCA plot of the task means learnt by our latent skill model. Each point represents the mean of a Gaussian that makes up our prior $p(\boldsymbol{z})$, with the colour denoting the nature of the task, and the label text denoting the domain of the dataset; ``unknown'' represents the mean of the extra $D+1$-th Gaussian that we use for unseen datasets. Note that all news domain datasets form a cluster, as do tasks with Reddit data.}
    \label{fig:task_mean_pca}
\vspace{-0.1cm}
\end{figure}


\paragraph{Results.}
Previous multitask natural language generation 
models often underperform
a single-task baseline trained on each dataset separately \citep{McCann:18,Radford:19}.
However, our results in Table~\ref{tab:best_results_summ} and Table~\ref{tab:multi_task} clearly demonstrate that 
our model (\textsc{Full}) which is trained on multiple tasks is the best performing model.
In terms of absolute scores (Table~\ref{tab:best_results_summ})
on each dataset and each metric (nine datasets, four summarisation metrics, one question answering metric),
multitask \textsc{Full} is the best model on 10 out of 27 cases.
The second best model (absolute scores) 
is a single task \textsc{NoLatent}, which is the best model in 7 cases.
Note that this single task model does not generalise to other tasks and
can only perform well on a specific dataset.

In terms of aggregated scores (Table~\ref{tab:multi_task}),
multitask \textsc{Full} again performs the best. The second best model under
this metric is the multitask \textsc{Base} model.
Overall results from both \textsc{NoDataset} and \textsc{NoLatent} show
that if either task information 
or the latent variable is removed, the multitask performance of 
the full model drops significantly.
This indicates that modelling latent skills in
a continuous shared space is
beneficial for performing multiple tasks with one model.

Comparing the results of task-specific and multitask models
for question answering datasets, it is evident that
multitask training significantly helps for question answering.
To evaluate whether this effect is simply down to the models sharing 
information among question answering tasks only, we train each model on only those tasks, 
and find worse performance compared to training on all tasks for every model---the best normalised
scores (comparable to numbers in Table~\ref{tab:multi_task}) we see in this condition are 87.08, 72.27 and 79.75 on MSMARCO, NewsQA and SQuAD respectively.
Our results show that
training a model on summarisation tasks can help improve question answering---\citet{Arumae:19} previously
observe the reverse direction of this phenomenon.

We visualise each of the Gaussian means from our prior $p(\boldsymbol{z})$ (projected into a two-dimensional space using PCA) learnt by our full model in Figure~\ref{fig:task_mean_pca}.
Our model appears to group datasets mainly by the domain the datasets come from, 
although the two non-news QA datasets also form a cluster, despite being from very different domains. The plot indicates that our model clusters tasks
that require similar skills in a meaningful way.

\subsection{Few-Shot Learning}
\label{sec:zero-shot}
    \begin{table}[t]
    \centering
    \resizebox{0.48\textwidth}{!}{
    \begin{tabular}{l lll}
    \toprule
    \multicolumn{1}{l}{\# examples} & 5 & 50 & 250 \\
    \midrule
    \multicolumn{4}{c}{News QA: zero-shot 32.44} \\
    \textsc{Infer} & 32.45 & 32.45 & 32.45 \\
    \textsc{Infer++} & 32.59 $\pm$1.36 & 33.00 $\pm$0.39 & 33.10 $\pm$0.15 \\
    \textsc{Gradient} & \textbf{32.98} $\pm$0.74 & \textbf{33.94} $\pm$1.22 & \textbf{33.52} $\pm$2.11 \\
    \midrule
    \multicolumn{4}{c}{NYT: zero-shot 29.04} \\
    \textsc{Infer} & 29.06 & 29.06 & 29.06 \\
    \textsc{Infer++} & \textbf{32.35} $\pm$0.93 & \textbf{32.64} $\pm$0.41 & 32.67 $\pm$0.28 \\
    \textsc{Gradient} & 29.36 $\pm$2.73 & 31.59 $\pm$1.71 & \textbf{33.55} $\pm$2.05 \\
    \midrule
    \multicolumn{4}{c}{TL;DR: zero-shot 11.38} \\
    \textsc{Infer} & \textbf{11.37} & \textbf{11.37} & \textbf{11.37} \\
    \textsc{Infer++} & 10.71 $\pm$1.02 & 10.94 $\pm$0.28 & 11.00 $\pm$0.13 \\
    \textsc{Gradient} & 11.15 $\pm$0.37 & 10.71 $\pm$0.51 & 10.32 $\pm$0.75 \\
    \bottomrule
    \end{tabular}
    }
    \caption{Few shot results for each adaptation method, including the standard deviations across hyperparameters. We report ROUGE-1 only for this experiment. Our best inference-based adaptation method achieves comparable or superior performance to gradient-based adaptation. In addition, our adaptation methods are more stable across hyperparameter values, as shown by their generally smaller standard deviations.
    For the TL;DR dataset, none of the adaptation methods is able to make use of the extra example to improve upon 
    the zero-shot performance.}
    \label{tab:few-shot}
\vspace{-0.3cm}
\end{table}

We next evaluate our model in the few-shot setting. In this evaluation,
the model is trained on a set of training tasks and then given a few
examples from a new task not seen at training. The model needs to adapt 
to the new task using these few-shot examples and is evaluated on test examples from the new task.

For the initial training stage, we train our model on 
all tasks other than the NYT, TL;DR and NewsQA datasets.
These held-out datasets are chosen 
to provide an example of a task similar to those in the 
training tasks (NYT), and two examples of datasets which require 
some generalisation (TL;DR and NewsQA). 
For the NewsQA dataset, the model has seen 
news domain summarisation data and out-of-domain QA data in
the initial training stage, and 
must generalise to news QA. 
For TL;DR, the model has 
seen Reddit dialogue data and out-of-domain summarisation data,
and must generalise to Reddit summarisation. The 
summarisation style of TL;DR is highly abstractive, 
which gives a harder adaptation task.

We use
the \textsc{Full} model which performs the best
in the previous experiment and compare three few-shot adaptation methods:
\begin{itemize}
    \item \textsc{Infer}: our basic inference-based adaptation (\S\ref{sec:finetuning}) that uses the output of the inference network directly.
    \item \textsc{Infer++}: our improved inference-based adaptation (\S\ref{sec:finetuning}) that refines its estimates further by optimising Eq.~\ref{eq:improved}. We use gradient descent to optimise Eq.~\ref{eq:improved} with learning rates in the set $\{0.1, 0.5, 1.0\}$ and maximum number of steps in the set $\{100, 200, 500\}$.
    \item \textsc{Gradient}: a baseline method that uses the fixed prior mean of $\vect{z}^{D+1}$ to generate each example, and fine-tunes all of the model parameters using gradient descent. We use learning rates in the set $\{1e^{-5}, 5e^{-5}, 1e^{-4}\}$ and steps per batch of examples in the set $\{1, 5, 10\}$.
\end{itemize}

We present batches of 5 few-shot training examples to our model, and evaluate our model after 5, 50 and 250 examples. 
We also show the zero-shot performance to provide an idea of whether the model can make
use of the few-shot examples to improve its performance.
As we do not have a separate development set in the few-shot scenario to tune hyperparameters, we average the test set performance across all hyperparameters over 5 different consecutive model checkpoints---to reduce the high variance of the gradient-based fine-tuning method---and report results in Table~\ref{tab:few-shot}.

Our results show that our best inference-based adaptation
method compares favourably to standard fine-tuning.
\textsc{Infer} does not provide any additional improvement over the zero-shot result. This suggests that the inference network does not generalise well to out-of-domain data. 
However, \textsc{Infer++} is able to improve over the zero-shot model 
performance (i.e., it is able to make use of the few-shot examples
to perform the task better) for 2 of the 3 held out datasets. The only dataset it does not improve on is the TL;DR summarisation corpus, which is a difficult adaptation task and no adaptation method works on this dataset.
Inference-based adaptation appears more stable to hyperparameter choices; the standard deviation of the inference-based methods is generally much lower than the gradient-based methods.

\section{Related Work}
\paragraph{Natural language generation.}
Traditional natural language generation approaches have focused 
on rule-based or template-based generators conditioning 
on a structured logical representation of the input \citep{Reiter:00}. With significant advances in deep learning, learning-based approaches 
conditioning on a wide range of input 
modalities have been explored \citep{Vinyals:15,Hinton:12}. 
In this paper, we focus on conditional language generation tasks 
where the input is also natural language. 
Such tasks include those in our training set and others (e.g., text simplification).
Many learning-based approaches have been 
employed to tackle this problem \citep{Kalchbrenner:13,Durrett:16,Zhang:17}. 

Our model is a latent variable conditional language model. Unconditional latent variable language models, such as those presented in \citet{Bowman:16}, have been shown to learn useful representations for tasks like discourse parsing \citep{Ji:16} and semi-supervised sentence classification \citep{Yang:17}. Further, they have been shown to learn smooth latent spaces that afford interpolation. Conditional latent variable language models have previously been used for open-domain dialogue \citep{Cao:17,Serban:17} and paraphrasing \citep{Narayan:16}.

\paragraph{Multitask learning for language.}
Multitask learning, \textit{sensu lato}, tries to improve the performance of a model on a particular task by leveraging the information contained in similar tasks \citep{Caruana:97}. \citet{Collobert:11} investigate multitask learning with neural models for NLP by sharing a feature extraction neural network across multiple different tasks. They show that powerful neural network feature extractors, trained using a language modelling objective on unlabelled data, can improve performance at specific natural language processing tasks such as part-of-speech tagging and named entity recognition. \texttt{word2vec} \citep{Mikolov:13} introduced a much simpler training objective for learning word representations, which showed promise when used in many downstream tasks. As computational power increased, methods such as ELMo \citep{Peters:18} and BERT \citep{Devlin:19}---which learn representations of words in context---currently hold the state-of-the-art on many natural language processing tasks.

For language generation specifically, both multitask learning \citep{Guo:18} and transfer learning \citep{Liu:19} have been investigated. A recent finding shows that large language models trained on a diverse collection of natural language data can generate coherent high-quality long-form text \citep{Radford:19}. However, it remains difficult to directly control the output of such a model for a particular task. Recent approaches include fine-tuning such models directly on a downstream task \citep{Gehrmann:19}, ensembling a large unconditional language model with a smaller auxiliary conditional model \citep{Dathathri:20}, or using it as the source model in a noisy channel decomposition \citep{Yee:19,Yu:19}.

\paragraph{Few-shot learning and skill modelling.}
Adapting a model to a new task using relatively few examples is a long standing goal of machine learning. One approach is to optimise the fine-tuning objective at training time, whether via gradient descent \citep{Andrychowicz:16,Finn:17} or a matching objective \citep{Vinyals:16}. Another approach is to treat few-shot learning as inference in a Bayesian model \citep{Gordon:19,Ravi:19}. \citet{Hausman:18} present a similar model to us in the context of learning transferrable skills for robotic control tasks. In addition, \citet{Garnelo:18} parametrise a model to directly estimate distributions given few-shot data. Recent work has considered using mixture-of-Gaussian latent variables in the context of continual unsupervised representation learning, where each component represents a cluster of related examples \citep{Rao:19}.

\section{Conclusion}
We present a generative model for multitask language generation which augments
an encoder-decoder model with a task embedding space for modelling latent skills.
We show that the resulting model can perform multiple language generation 
tasks simultaneously better
compared to models which do not use task information or only learns a pointwise task embedding. 
We also show that our model can generalise to unseen tasks with few-shot examples by 
inference and adaptation in the latent space, and that this inference procedure is competitive with a standard fine-tuning method that adapts all model parameters
in terms of performance and is more stable across hyperparameter choices.

The main limitations of our model are that we need to fix the number of tasks in advance and need to observe the dataset identifier.
Exciting avenues for future work include adding the ability
to continually grow our model (e.g., by assuming a Dirichlet process prior over the tasks) and designing better ways to incorporate unlabelled data.


\section*{Acknowledgements}
We thank Angeliki Lazaridou for helpful comments on an
earlier draft of this paper and the Language group at DeepMind
for valuable discussions.

\bibliographystyle{icml2020}
\bibliography{example_paper_short}

\appendix

\section{Complete Results}
\label{app:full_results}

We show the complete results on each dataset for \textsc{Full}, \textsc{NoDataset}, \textsc{NoLatent}, and \textsc{Base} in Table~\ref{tab:full_model_full_results}, Table~\ref{tab:no_dataset_full_results}, Table~\ref{tab:no_latent_full_results}, and Table~\ref{tab:base_model_full_results} respectively.

\begin{table*}[ht]
    \centering
    \small
    \begin{tabular}{l ccccc ccccc}
        \toprule
         & \multicolumn{5}{c}{Single task} & \multicolumn{5}{c}{Multitask} \\
         \cmidrule(r){2-6} \cmidrule(l){7-11}
        Dataset & R1 & R2 & RL & BLEU & F1 & R1 & R2 & RL & BLEU & F1 \\
        \midrule 
        Gigaword & 46.95 & 23.50 & 44.01 & 19.71 & -- & 45.80 & 22.56 & 43.15 & 19.20 & -- \\
        CNN/DM & 41.53 & 16.84 & 27.87 & 12.96 & -- & 40.52 & 16.29 & 28.01 & 13.68 & -- \\
        Newsroom & 28.62 & 18.99 & 26.15 & 35.82 & -- & 32.76 & 23.10 & 30.09 & 34.35 & -- \\
        NYT & 44.27 & 27.88 & 36.52 & 29.67 & -- & 44.79 & 28.32 & 36.65 & 30.01 & -- \\
        TL;DR & 14.27 & 2.14 & 10.54 & 2.18 & -- & 13.43 & 1.72 & 9.98 & 1.76 & -- \\
        Wikihow & 26.00 & 6.33 & 18.39 & 6.41 & -- & 26.73 & 8.13 & 20.48 & 7.28 & -- \\
        \midrule
        MSMARCO & -- & -- & -- & -- & 65.46 & -- & -- & -- & -- & 61.90 \\
        NewsQA & -- & -- & -- & -- & 11.06 & -- & -- & -- & -- & 48.69 \\
        SQuAD & -- & -- & -- & -- & 13.71 & -- & -- & -- & -- & 73.81 \\
        \bottomrule
    \end{tabular}
    \caption{Results for the \textsc{Full} model across all metrics and all tasks in the single and multitask conditions.}
    \label{tab:full_model_full_results}
\end{table*}

\begin{table*}[ht]
    \centering
    \small
    \begin{tabular}{l ccccc ccccc}
        \toprule
         & \multicolumn{5}{c}{Single task} & \multicolumn{5}{c}{Multitask} \\
         \cmidrule(r){2-6} \cmidrule(l){7-11}
        Dataset & R1 & R2 & RL & BLEU & F1 & R1 & R2 & RL & BLEU & F1 \\
        \midrule 
        Gigaword & 48.59 & 24.73 & 45.74 & 21.51 & -- & 40.96 & 18.28 & 38.79 & 14.93 & -- \\
        CNN/DM & 41.82 & 16.72 & 28.23 & 13.63 & -- & 40.83 & 14.52 & 26.37 & 10.99 & -- \\
        Newsroom & 33.35 & 24.19 & 30.61 & 39.45 & -- & 29.67 & 19.91 & 26.65 & 19.01 & -- \\
        NYT & 44.64 & 27.44 & 36.12 & 29.64  & -- & 40.25 & 22.64 & 31.32 & 22.93 & -- \\
        TL;DR & 13.99 & 2.01 & 10.36 & 1.91 & -- & 13.12 & 1.42 & 9.17 & 1.02 & -- \\
        Wikihow & 25.64 & 6.49 & 18.86 & 6.27 & -- & 25.88 & 6.10 & 17.68 & 5.99 & -- \\
        \midrule
        MSMARCO & -- & -- & -- & -- & 59.86 & -- & -- & -- & -- & 54.07 \\
        NewsQA & -- & -- & -- & -- & 12.52 & -- & -- & -- & -- & 42.48 \\
        SQuAD & -- & -- & -- & -- & 13.22 & -- & -- & -- & -- & 66.18 \\
        \bottomrule
    \end{tabular}
    \caption{Results for the \textsc{NoDataset} model across all metrics and all tasks in the single and multitask conditions.}
    \label{tab:no_dataset_full_results}
\end{table*}

\begin{table*}[ht]
    \centering
    \small
    \begin{tabular}{l ccccc ccccc}
        \toprule
         & \multicolumn{5}{c}{Single task} & \multicolumn{5}{c}{Multitask} \\
         \cmidrule(r){2-6} \cmidrule(l){7-11}
        Dataset & R1 & R2 & RL & BLEU & F1 & R1 & R2 & RL & BLEU & F1 \\
        \midrule 
        Gigaword & 50.14 & 26.56 & 47.28 & 23.20 & -- & 47.09 & 23.77 & 44.37 & 20.14 & -- \\
        CNN/DM & 37.83 & 15.05 & 25.67 & 11.00 & -- & 36.45 & 14.69 & 25.39 & 10.51 & -- \\
        Newsroom & 34.35 & 24.50 & 31.90 & 36.17 & -- & 31.09 & 22.20 & 28.85 & 19.85 & -- \\
        NYT & 42.38 & 26.52 & 34.56 & 24.68 & -- & 43.40 & 27.62 & 35.94 & 25.92 & -- \\
        TL;DR & 11.13 & 1.93 & 9.45 & 1.62 & -- & 9.76 & 1.44 & 8.29 & 0.69 & -- \\
        Wikihow & 17.97 & 4.47 & 13.63 & 3.27 & -- & 18.51 & 6.25 & 15.69 & 2.41 & --\\
        \midrule
        MSMARCO & -- & -- & -- & -- & 51.98 & -- & -- & -- & -- & 61.89 \\
        NewsQA & -- & -- & -- & -- & 8.86 & -- & -- & -- & -- & 49.00 \\
        SQuAD & -- & -- & -- & -- & 9.13 & -- & -- & -- & -- & 73.96 \\
        \bottomrule
    \end{tabular}
    \caption{Results for the \textsc{NoLatent} model across all metrics and all tasks in the single and multitask conditions.}
    \label{tab:no_latent_full_results}
\end{table*}

\begin{table*}[ht]
    \centering
    \small
    \begin{tabular}{l ccccc ccccc}
        \toprule
         & \multicolumn{5}{c}{Single task} & \multicolumn{5}{c}{Multitask} \\
         \cmidrule(r){2-6} \cmidrule(l){7-11}
        Dataset & R1 & R2 & RL & BLEU & F1 & R1 & R2 & RL & BLEU & F1 \\
        \midrule 
        Gigaword & 49.96 & 26.32 & 47.14 & 22.91 & -- & 46.92 & 23.72 & 44.32 & 20.23 & -- \\
        CNN/DM & 37.54 & 14.83 & 25.50 & 10.76 & -- & 38.13 & 15.52 & 26.54 & 12.13 & --\\
        Newsroom & 30.09 & 21.15 & 27.79 & 21.88 & -- & 30.93 & 21.64 & 28.55 & 24.27 & -- \\
        NYT & 42.73 & 26.41 & 34.84 & 26.60 & -- & 43.90 & 27.96 & 36.25 & 27.42 & -- \\
        TL;DR & 9.07 & 1.34 & 7.75 & 0.70 & -- & 10.92 & 1.63 & 9.00 & 1.45 & --\\
        Wikihow & 20.70 & 5.11 & 15.42 & 4.00 & -- & 17.88 & 5.93 & 15.13 & 2.85 & --\\
        \midrule
        MSMARCO & -- & -- & -- & -- & 55.46 & -- & -- & -- & -- & 64.64 \\
        NewsQA & -- & -- & -- & -- & 10.50 & -- & -- & -- & -- & 48.37 \\
        SQuAD & -- & -- & -- & -- & 9.61 & -- & -- & -- & -- & 73.26 \\
        \bottomrule
    \end{tabular}
    \caption{Results for the \textsc{Base} model across all metrics and all tasks in the single and multitask conditions.}
    \label{tab:base_model_full_results}
\end{table*}


\section{Summarisation Style Transfer}
\label{app:styletransfer}

In this section, we report the results of experiment of summarisation style transfer.
We use the \textsc{Full} model and consider two evaluation setups in this experiment:
\begin{itemize}
\item We take articles from the NYT development set and compare reference summaries, summaries generated from the prior mean corresponding to the NYT dataset $\vect{z}^{\text{NYT}}$, and summaries generated from the prior mean corresponding to the Newsroom dataset $\vect{z}^{\text{Newsroom}}$. We show representative samples in Table \ref{tab:nyt_1} and Table \ref{tab:nyt_2}.
\item We take articles from the Newsroom development set and compare the same selection of summaries as above. We show representative samples in Table \ref{tab:newsroom_1} and Table \ref{tab:newsroom_2}.
\end{itemize}

Inspecting the samples, we can observe that the model has learnt different summarisation styles as a result of the different training data. 
The summaries generated using $\vect{z}^{\text{Newsroom}}$
often seem to consist of the article lede, whereas summaries generated with $\vect{z}^{\text{NYT}}$ consist of extracted phrases which are more evenly distributed throughout the article. 
We note that while the summaries generated using a skill
representation that is not intended for that dataset
(i.e., $\vect{z}^{\text{Newsroom}}$ for NYT and vice versa)
score less using standard metrics compared 
to the summaries generated using the ``correct'' skill representation 
(35.9 R1 vs. 44.27 R1 on NYT and 28.1 R1 vs. 28.62 on Newsroom), 
the resulting summaries are still valid summaries as indicated by
the reasonably high scores.
This suggests that our model learns useful summarisation skills
that can be generalized to
other domains.

\begin{table*}[htb]
    \centering
    \begin{tabular}{l l}
    \toprule
        Article & \pbox{0.7\textwidth}{WHEN people die in train collisions, like the 2 engineers and a passenger killed in a commuter train crash in Secaucus, N.J., or the 11 on a Maryland commuter train that hit an Amtrak train in Silver Spring, both in early 1996, there are national headlines and detailed safety investigations. But the attention might be better directed elsewhere: Far more people, more than 400 last year, died in the less noted but far more numerous accidents in which trains collided not with other trains, but with motor vehicles. The Federal Railroad Administration counted more than 4,000 car-train collisions last year. ''Something comes over the fax every minute of every day,'' David Bolger, a spokesman for the agency, said with only slight exaggeration. His agency is running a long-term campaign to reduce the toll. The Transportation Department's goal, set in 1991, is to reduce by 25 percent by 2001 the number of places where tracks cross roads. So far the number is down by a little more than 6 percent, to 268,000, with three years to go. The theory is that if crossings are combined, towns or highway departments can afford to pay for improvements at the ones that survive. Those costs are high. The Transportation Department says a simple ''crossbuck ...} \\
        \midrule
        Reference summary & \pbox{0.7\textwidth}{Matthew L Wald comments on growing concern about train-car collisions, noting that Federal Railroad Administration counted more than 4,000 car-train collisions last year; Transportation Dept goal, set in 1991, is to reduce by 25 percnet [\textit{sic}] by 2001 number of places where tracks cross roads; so far, that number is down by little more than 6 percent, to 268,000; photo (M)} \\
        \midrule
        NYT summary style & \pbox{0.7\textwidth}{Federal Railroad Administration has counted more than 4,000 car-train collisions in 1996, with more than 400 deaths in less noted but far more numerous accidents in which trains collided not with other trains, but with motor vehicles; Transportation Department's goal is to reduce by 25 percent by 2001 number of places where tracks cross roads; theory is that if crossings are combined, towns or highway departments can afford to pay for improvements at ones that survive; those costs are high; photo (Special section, Cars) (M)} \\
        \midrule
        Newsroom summary style & \pbox{0.7\textwidth}{WHEN people die in train collisions, like the 2 engineers and a passenger killed in a commuter train crash in Secaucus, N.J., or the 11 on a Maryland commuter train that hit an Amtrak train in Silver Spring, both in early 1996, there are national headlines and detailed safety investigations. But the attention might be better directed elsewhere: Far more people, more than 400 last year, died in the less noted but far more numerous accidents in which trains collided not with other trains, but with motor vehicles. The Federal Railroad Administration counted more than 4,000 car-train collisions last year. ''Something comes over the fax every minute of every day,'' David Bolger, a spokesman for the agency, said with only slight exaggeration.} \\
    \bottomrule
    \end{tabular}
    \caption{An NYT article with a reference summary, and two model-generated summaries using $\vect{z}^{\text{NYT}}$ and $\vect{z}^{\text{Newsroom}}$.}
    \label{tab:nyt_1}
\end{table*}

\begin{table*}[htb]
    \centering
    \begin{tabular}{l l}
    \toprule
        Article & \pbox{0.7\textwidth}{Months ago, Alaska posted a list of its 1,600 paroled sex offenders on the World Wide Web. California provides information on about 64,000 offenders on CD-ROM at any county sheriff's office or big-city police department. And three months ago, Connecticut began letting local police stations release the names, addresses and pictures of paroled offenders to anyone who calls. But in New Jersey, the state that inspired the nationwide movement for public warnings about the presence of sex offenders, information is held far more tightly. Since New Jersey's community-notification law was hurriedly enacted in 1994 amid the outcry that followed the killing of Megan Kanka of Hamilton Township, it has faced an onslaught of constitutional challenges, and nearly three years of review and revision by Federal and state judges. Indeed, the original ''Megan's Law'' is soon to emerge from legal limbo as one of the most restricted in the nation, with tight limits on which offenders are identified, who has access to the information and how those facts may be passed on. By the end of this month, when most of New Jersey's 21 county prosecutors are expected to begin issuing names and addresses of offenders, only about 600 parolees will be listed, those classified as posing a moderate or high risk of committing new ...} \\
        \midrule
        Reference summary & \pbox{0.7\textwidth}{New Jersey, which inspired other states to make known the names of paroled sex offenders, has faced onslaught of constitutional challenges to its own Megan's law, and now the law is emerging from nearly three years of judicial review and revision as one of most restricted in nation; it sets tight limits on which offenders are identified, who has access to information and how those facts may be passed on (M)} \\
        \midrule
        NYT summary style & \pbox{0.7\textwidth}{New Jersey, which inspired nationwide movement for public warnings about presence of sex offenders, has faced onslaught of constitutional challenges, and nearly three years of review and revision by Federal and state judges; original 'Megan's Law' is soon to emerge from legal limbo as one of most restricted in nation, with tight limits on which offenders are identified, who has access to information and how those facts may be passed on; photo (M)} \\
        \midrule
        Newsroom summary style & \pbox{0.7\textwidth}{Months ago, Alaska posted a list of its 1,600 paroled sex offenders on the World Wide Web. California provides information on about 64,000 offenders on CD-ROM at any county sheriff's office or big-city police department. And three months ago, Connecticut began letting local police stations release the names, addresses and pictures of paroled offenders to anyone who calls. But in New Jersey, the state that inspired the nationwide movement for public warnings about the presence of sex offenders, information is held far more tightly.} \\
    \bottomrule
    \end{tabular}
    \caption{An NYT article with a reference summary, and two model-generated summaries using $\vect{z}^{\text{NYT}}$ and $\vect{z}^{\text{Newsroom}}$.}
    \label{tab:nyt_2}
\end{table*}

\begin{table*}[htb]
    \centering
    \begin{tabular}{l l}
    \toprule
        Article & \pbox{0.7\textwidth}{On the road ... Pilgrim and Ambassador, 1977, by Raghubir Singh. See it in Everything Was Moving at the Barbican, London August is a quiet month for photography shows, so here's a preview of some of the exhibition highlights for the next few months. The most anticipated London show is surely Tate Modern's ambitious double header William Klein/Daido Moriyama, which opens on 10 October. Taking the cities of New York and Tokyo as its starting point, the show contrasts the approaches of two pioneers of impressionistic urban photography. It considers the influence of Klein's seminal 1956 book, Life Is Good and Good for You in New York, on Japanese photography, and Moriyama in particular. The prodigiously productive Moriyama was a founder of the radical Provoke movement in Japan and, alongside previously unseen vintage prints, the exhibition explores photography's role in the representation of protest movements and civil unrest. This is an ambitious show that will be a chance for many of us to see lots of Moriyama's images outside of book form for the first time. I, for one, cannot wait. The other big London exhibition is the Barbican's group show, Everything Was Moving: Photography from the 60s ...} \\
        \midrule
        Reference summary & \pbox{0.7\textwidth}{From an ambitious survey of 1960s-70s photography in London to Kohei Yoshiyuki's controversial work in Liverpool and Amsterdam's Unseen Photo Fair, there's a lot to see. By Sean O'Hagan} \\
        \midrule
        NYT summary style & \pbox{0.7\textwidth}{Barbican, London August, previews Tate Modern's ambitious double header William Klein/Daido Moriyama, which opens on 10 October; show contrasts approaches of two pioneers of impressionistic urban photography; it considers influence of Klein's seminal 1956 book, Life Is Good and Good for You in New York, on Japanese photography, and Moriyama in particular; other big London exhibition is Barbican's group show, Everything Was Moving: Photography from 60s; photos (M)} \\
        \midrule
        Newsroom summary style & \pbox{0.7\textwidth}{Two pioneers of impressionistic urban photography are the latest in London's show, Everything Was Moving: Photography from the 60s, says Jonathan Jones!!!} \\
    \bottomrule
    \end{tabular}
    \caption{A Newsroom article with a reference summary, and two model-generated summaries using $\vect{z}^{\text{NYT}}$ and $\vect{z}^{\text{Newsroom}}$.}
    \label{tab:newsroom_1}
\end{table*}

\begin{table*}[htb]
    \centering
    \begin{tabular}{l l}
    \toprule
        Article & \pbox{0.7\textwidth}{BANGALORE, India, June 4 The world's biggest computer services company could not have chosen a more appropriate setting to lay out its strategy for staying on top. A building housing I.B.M.'s software laboratory and application service teams on the company's corporate campus in Bangalore, India. On Tuesday, on the expansive grounds of the Bangalore Palace, a colonial-era mansion once inhabited by a maharajah, the chairman and chief executive of I.B.M., Samuel J. Palmisano, will address 10,000 Indian employees. He will share the stage with A. P. J. Abdul Kalam, India's president, and Sunil Mittal, chairman of the country's largest cellular services provider, Bharti Tele-Ventures. An additional 6,500 employees will look in on the town hall-style meeting by satellite from other Indian cities. On the same day, Mr. Palmisano and other top executives will meet here with investment analysts and local customers to showcase I.B.M.'s global integration capabilities in a briefing customarily held in New York. During the week, the company will lead the 50 analysts on a tour of its Indian operations. The meetings are more than an exercise in public and investor relations. They ...} \\
        \midrule
        Reference summary & \pbox{0.7\textwidth}{India provides I.B.M. with its fastest-growing market and a crucial base for delivering services to much of the world.}\\
        \midrule
        NYT summary style & \pbox{0.7\textwidth}{IBM chairman and chief executive Samuel J Palmisano will address 10,000 Indian employees on expansive grounds of Bangalore Palace, colonial-era mansion once inhabited by maharajah, will meet with investment analysts and local customers to showcase IBM's global integration capabilities in briefing customarily held in New York; during week, company will lead 50 analysts on tour of its Indian operations; meetings are more than exercise in public and investor relations; photo (M)} \\
        \midrule
        Newsroom summary style & \pbox{0.7\textwidth}{The world's biggest computer services company could not have chosen a more appropriate setting to lay out its strategy for staying on top.} \\
    \bottomrule
    \end{tabular}
    \caption{A Newsroom article with a reference summary, and two model-generated summaries using $\vect{z}^{\text{NYT}}$ and $\vect{z}^{\text{Newsroom}}$.}
    \label{tab:newsroom_2}
\end{table*}

\end{document}